\documentclass[11pt]{article}
\usepackage{acl2015}
\usepackage{times}
\usepackage{url}
\usepackage{latexsym}
\usepackage{amsmath}
\usepackage{amstext}
\usepackage{amssymb}
\usepackage{graphicx}
\usepackage{color}
\usepackage{subcaption}
\usepackage{enumitem}
\usepackage{bm}
\usepackage{multirow}

\newcommand{\bX}{{\bf x}}
\newcommand{\bY}{{\bf y}}
\newcommand{\bZ}{{\bf z}}
\newcommand{\bF}{{\bf f}}
\newcommand{\bW}{{\bf W}}

\title{Paraphrase-Supervised Models of Compositionality}

\author{Avneesh Saluja \and Chris Dyer \\
	Carnegie Mellon University \\
  Pittsburgh, PA, 15213, USA \\
  {\tt \{avneesh,cdyer\}@cs.cmu.edu} \\\And
  Jean-David Ruvini \\
  eBay Inc.\\
  San Jose, CA, 95125, USA \\
  {\tt jean-david.ruvini@ebay.com} \\}

\date{}

\begin{document}
\maketitle
\begin{abstract}
Compositional vector space models of meaning promise new solutions to stubborn language understanding problems. 
This paper makes two contributions toward this end: (i) it uses automatically-extracted paraphrase examples as a source of supervision for training compositional models, replacing previous work which relied on manual annotations used for the same purpose, and (ii) develops a context-aware model for scoring phrasal compositionality. 
	Experimental results indicate that these multiple sources of information can be used to learn partial semantic supervision that matches previous techniques in intrinsic evaluation tasks. 
	Our approaches are also evaluated for their impact on a machine translation system where we show improvements in translation quality, demonstrating that compositionality in interpretation correlates with compositionality in translation. 
 \end{abstract}

\section{Introduction}

Numerous lexical semantic properties are captured by representations encoding distributional properties of words, as has been demonstrated in a variety of tasks \cite{Turian2010,Turney2010,Mikolov2013b}.
However, this distributional account of meaning does not scale to larger units like phrases and sentences \cite[\emph{inter alia}]{Sahlgren2006,Collobert2011},\footnote{There are simply far more distinct phrasal units whose representations have to be learned from the same amount of data.} motivating research into compositional models that \emph{combine} word representations to produce representations of the semantics of longer units \cite{Mitchell2010,Baroni2010,Socher2013}.  
Previous work has learned these models using autoencoder formulations \cite{Socher2011} or limited human supervision \cite{Mitchell2010}.
In this work, we explore the hypothesis that the equivalent knowledge about how words compose can be obtained through monolingual paraphrases that have been extracted using word alignments and an intermediate language \cite{Ganitkevich2013}.
Confirming this hypothesis would allow the rapid development of compositional models in a large number of languages. 

As their name suggests, these models also impose the assumption that longer units like phrases are \textbf{compositional}, i.e., a phrase's meaning can be understood from the literal meaning of its parts. 
However, countless examples that run contrary to the assumption exist, and handling these \textbf{non-compositional} phrases has been problematic and of long-standing interest in the community \cite{Lin1999,Sag2002}.
(Non-) Compositionality detection can provide vital information to other language processing systems on whether a multiword unit should be treated semantically as a single entity or not, and scoring this phenomenon is particularly relevant for downstream tasks like machine translation (MT) or information retrieval. 
We explore the hypothesis that contextual evidence can be used to determine the relative degree to which a phrase is meant compositionally.  


Rather than focusing purely on intrinsic clean-room evaluations, the goal of this work is to learn relatively accurate context-sensitive compositional models that are also directly applicable in real-world, noisy-data scenarios. 
This objective necessitates certain design decisions, and to this end we propose a robust, scalable framework that learns compositional functions and scores relative phrasal compositionality.  
We make three contributions: first, a novel way to learn compositional functions for part-of-speech pairs that uses supervision from an automatically-extracted list of paraphrases (\S\ref{sec:ppdb}). 
Second, a context-dependent scoring model that scores the relative compositionality of a phrase \cite{McCarthy2003} by computing the likelihood of its context given its paraphrase-learned representation (\S\ref{sec:scoring}). 
And third, an evaluation of the impact of compositionality knowledge in an end-to-end MT setup.
Our experiments (\S\ref{sec:experiments}) reveal that using supervision from automatically extracted paraphrases produces compositional functions with equivalent performance to previous approaches that have relied on hand-annotated training data. 
Furthermore, compositionality features consistently improve the translations produced by a strong English--Spanish translation system.

\section{Parametric Composition Functions}
\label{sec:learning}

We formalize composition as a function $\bF (\bX, \bY)$ that maps $N$-dimensional vector representations of phrase constituents $\bX, \bY \in \mathbb{R}^{N \times 1}$  to an $N$-dimensional vector representation of the phrase\footnote{We discuss handling phrases longer than 2 words in \S\ref{sec:longer-phrases}.}, i.e., the \emph{composed} representation. 
A phrase is defined as any contiguous sequence of words of length 2 or greater, and does not have to adhere to constituents in a phrase structure grammar. 
This definition is in line with our MT application and ignores ``gappy'' noncontiguous phrases, but this pragmatic choice does exclude many verb-object relations \cite{Koehn2003}.  
We assume the existence of word-level vector representations for every word in our vocabulary of size $V$.
Compositionality is modeled as a bilinear map, and two classes of linear models with different levels of parametrization are proposed.  
Unlike previous work \cite[\emph{inter alia}]{Baroni2010,Socher2013,Grefenstette2013} where the functions are  word-specific, our compositional functions operate on part-of-speech (POS) tag pairs, which facilitates learning by drastically reducing the number of parameters, and only requires a shallow syntactic parse of the input. 

\subsection{Concatenation Models}
\label{sec:concat}

Our first class of models is a generalization of the additive models introduced in \newcite{Mitchell2008}:
\begin{align}
	\bF (\bX, \bY) &= \bW [\bX;\bY]
	\label{eq:concat}
\end{align}
where the notation $[\bX; \bY]$ represents a vertical (row-wise) concatenation of two vectors; namely, the concatenation that results in a $2N \times 1$-sized vector.  
In addition to the $N \times V$ parameters for the word vector representations that are provided \emph{a priori}, this model introduces $N \times 2N \times T$ parameters, where $T$ is the number of POS-tag pairs we consider.  

\newcite{Mitchell2008} significantly simplify parameter estimation by assuming a certain structure for the parameter matrix $\bW$, which is necessary given the limited human-annotated data they use.   
For example, by assuming a block-diagonal structure, we get a scaled element-wise addition model $f_i (x_i, y_i) = \alpha_i x_i + \beta_i y_i$. 
While not strictly in this category due to the non-linearities involved, neural network-based compositional models \cite{Socher2013,Hermann2013} can be viewed as concatenation models, although the order of concatenation and matrix multiplication is switched. 
However, these models introduce more than $V \times N^2$ parameters. 

\subsection{Tensor Models}
\label{sec:tensor}

The second class of models leverages pairwise multiplicative interactions between the components of the two word vectors:
\begin{align}
	\bF (\bX, \bY) &= (\bW \times_3 \bY) \times_2 \bX
	\label{eq:tensor}
\end{align}
where $\times_n$ corresponds to a tensor contraction along the $n\textsuperscript{th}$ mode of the tensor $\bW$. 
In this case, we first compute a contraction (tensor-vector product) between $\bW$ and $\bY$ along $\bW$'s third mode, corresponding to interactions with the second word vector of a two-word phrase and resulting in a matrix, which is then multiplied along its second mode (corresponding to traditional matrix multiplication on the right) by $\bX$.  
The final result is an $N \times 1$ vector.  
This model introduces $N \times N \times N \times T$ parameters.  

Tensor models are a generalization of the element-wise multiplicative model \cite{Mitchell2008}, which permits non-zero values only on the tensor diagonal.  
Operating at the vocabulary level, the model of \newcite{Baroni2010} has interesting parallels to our tensor model. 
They focus on adjective--noun relationships and learn a specific matrix for every adjective in their dataset; in our case, the specific matrix for each adjective has a particular form, namely that it can be factorized into the product of a tensor and a vector; the tensor corresponds to the actual adjective--noun combiner function, and the vector corresponds to specific lexical information that the adjective carries. 
This concept generalizes to other POS pairs: for example, multiplying the tensor that represents determiner-noun combinations along the second mode with the vector for ``the'' results in a matrix that represents the semantic operation of definiteness. 
Learning these parameters jointly is statistically more efficient than separately learning versions for each word.

\subsection{Longer Phrases}
\label{sec:longer-phrases}

The proposed models operate on pairs of words at a time. 
To handle phrases of length greater than two, we greedily construct a left-branching tree of the phrase constituents that eventually dictates the application of the learned bilinear maps.\footnote{We also tried constructing right-branching trees, but found that performance was never as good as the left-branching ones.}
For each internal tree node, we consider the POS tags of its children: if the right child is a noun, and the left child is either a noun, adjective, or determiner, then the internal node is marked as a noun, otherwise we mark it with a generic \emph{other} tag. 
At the end of the procedure, unattached nodes (words) are attached at the highest point in the tree. 

After the tree is constructed, we can compute the overall phrasal representation in a bottom-up manner, guided by the labels of leaf and internal nodes.
We note that the emphasis of this work is not to compute sentence-level representations. 
This goal has been explored in recent research \cite{Le2014,Kalchbrenner2014}, and combining our models with methods presented therein for sentence-level representations is straightforward.  

\section{Learning}
\label{sec:ppdb}
The models described above rely on parameters $\bW$ that must be learned. 
In this section, we argue that automatically constructed databases of paraphrases provide adequate supervision for learning notions of compositionality. 

\subsection{Supervision from Automatic Paraphrases}

The Paraphrase Database \cite[PPDB]{Ganitkevich2013} is a collection of ranked monolingual paraphrases that have been extracted from word-aligned parallel corpora using the bilingual pivot method \cite{Bannard2005}. 
The underlying assumption is that if two strings in the same language align to the same string in another language, then the strings in the original language share the same meaning. 
Paraphrases are ranked by their word alignment scores, and in this work we use the preselected \textsc{small} portion of PPDB as our training data.
Although we can directly extract phrasal representations of a pre-specified list of phrases from the corpus used to compute word representations \cite{Baroni2010}, this approach is both computationally and statistically inefficient: the number of phrases increases exponentially in the length of the phrase, and correspondingly the occurrence of any individual phrase decreases exponentially. 
We can thus circumvent these computational and statistical issues by using monolingual paraphrases.  

The training data is filtered to provide only two-to-one word paraphrase mappings, and the multiword portion of the paraphrase is subsequently POS-tagged.
Table \ref{tab:pos-stats} provides a breakdown of such paraphrases by their POS pair type.  
Given the lack of context when tagging, it is likely that the POS tagger yields the most probable tag for words and not the most probable tag given the (limited) context. 
Furthermore, even the higher quality portions of PPDB yield paraphrases of ranging quality, ranging from non-trivial mappings such as \emph{young people} $\rightarrow$ \emph{youth}, to redundant ones like \emph{the ceasefire} $\rightarrow$ \emph{ceasefire}. 
However, PPDB-like resources are more easily available than human-annotated resources (in multiple languages too: \newcite{Ganitkevich2014}), so it is imperative that methods which learn compositional functions from such sources handle noisy supervision adequately. 

\begin{table}[h!]
  \begin{center}
    \begin{tabular}{p{0.25\linewidth}r}
      \hline
      POS Pair & Size \\
	  \hline
      DT--NN & 10,982 \\
	  NN--NN &  4781 \\
	  JJ--NN & 3924 \\
  	  VB--VB  &  2021 \\
      RB--JJ &  1640 \\
	  \emph{other}  & 8548 \\
	\end{tabular}
  \end{center}
  \caption{Number of paraphrase examples per POS pair type out of the two-to-one word paraphrases in the \textsc{small} version of PPDB (using the Penn Treebank tag-set). We distinguish between the five most common POS pair types, and group the remaining pairs into the generic \emph{other} category.}
  \label{tab:pos-stats}
\end{table}

\subsection{Parameter Estimation}

\begin{figure*}[t!]
	\begin{center}
	\begin{subfigure}{\columnwidth}
		\centering
		\includegraphics[width=\columnwidth,keepaspectratio=true]{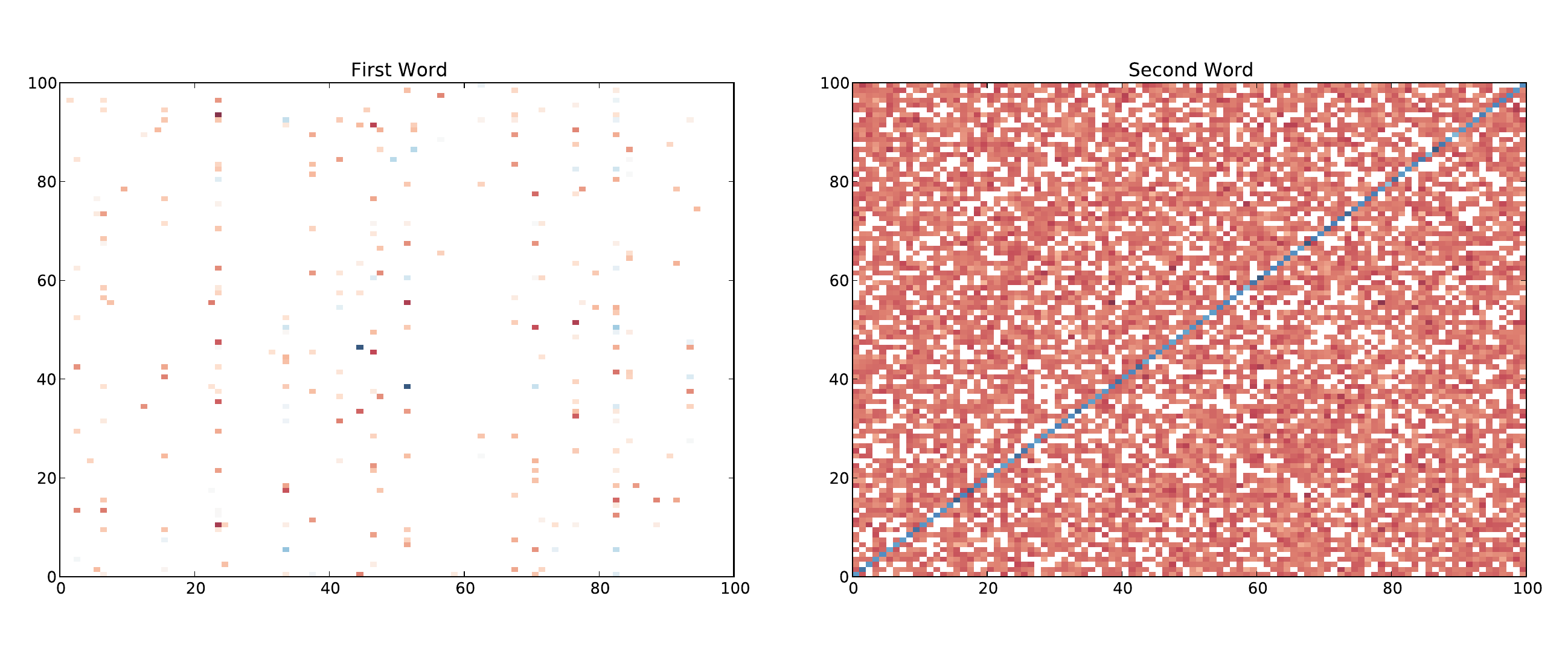}	
		\caption{\small DT--NN}
		\label{fig:dt_nn}			
	\end{subfigure}
	\begin{subfigure}{\columnwidth}
		\centering
		\includegraphics[width=\columnwidth,keepaspectratio=true]{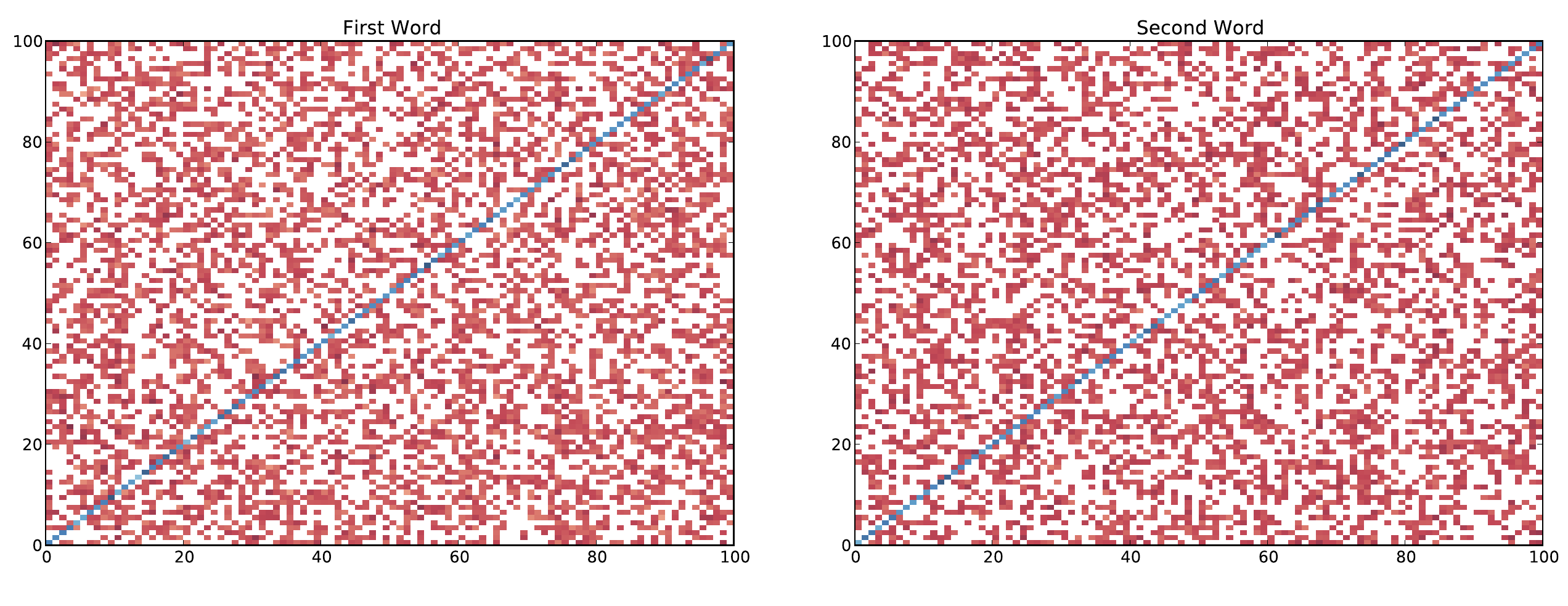}		
		\caption {\small NN--NN}
		\label{fig:nn_nn}
	\end{subfigure}
	\end{center}
	\begin{center}
	\begin{subfigure}{\columnwidth}
		\centering
		\includegraphics[width=\columnwidth,keepaspectratio=true]{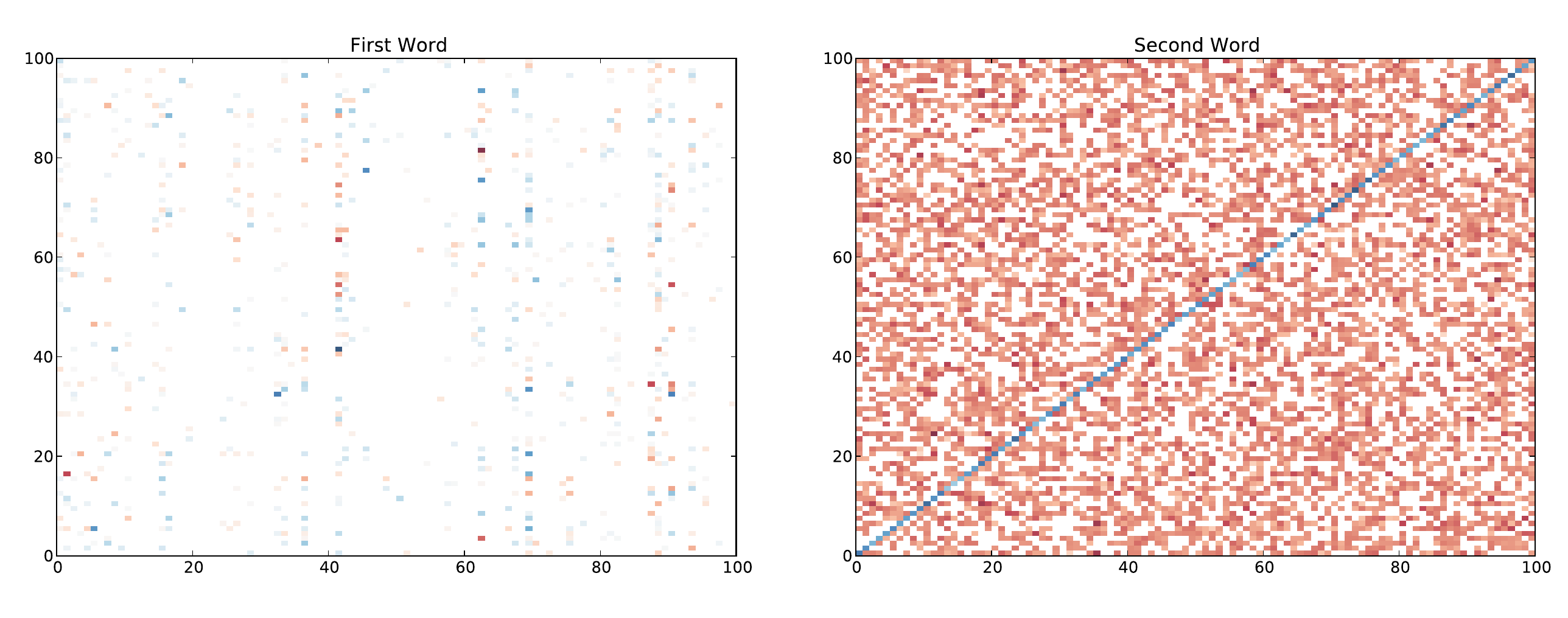}		
		\caption {\small VB--VB}
		\label{fig:vb_vb}
	\end{subfigure}
	\begin{subfigure}{\columnwidth}
		\centering
		\includegraphics[width=0.95\columnwidth,keepaspectratio=true]{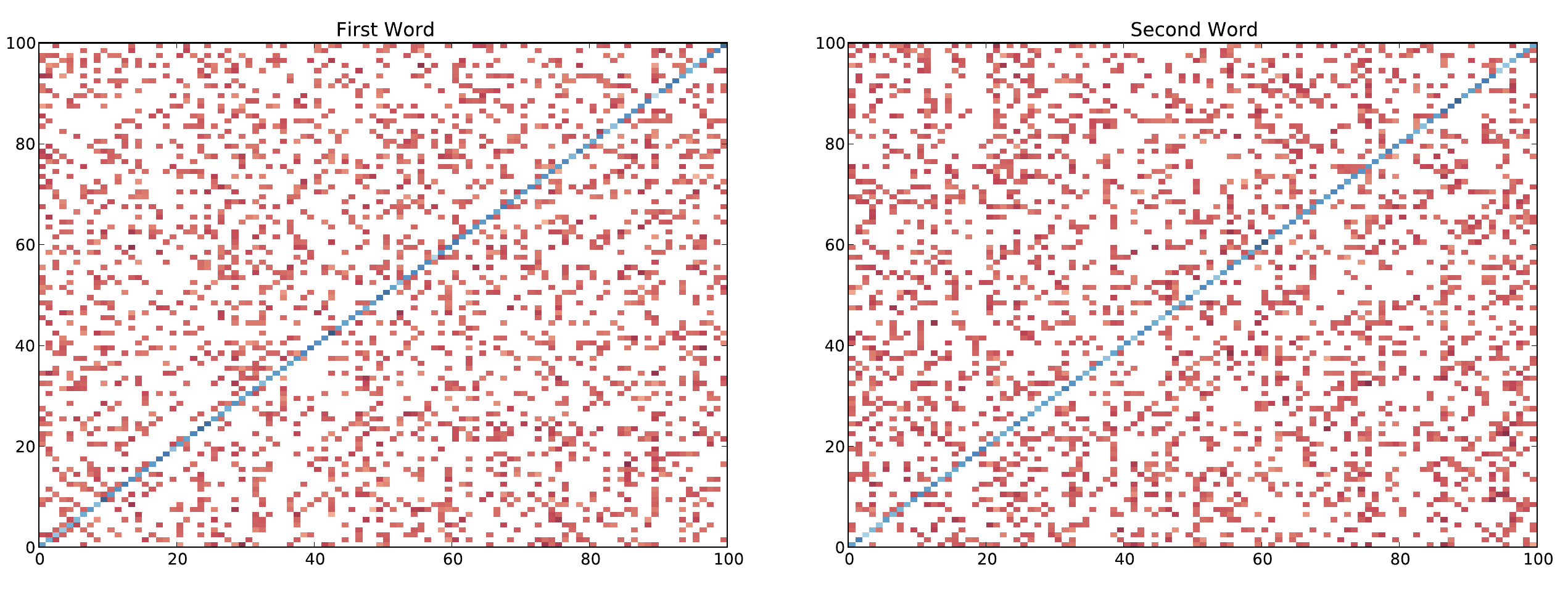}	
		\caption{\small \emph{other}}
		\label{fig:x_x}			
	\end{subfigure}
	\end{center}	
	\caption{Parameter heat-maps for specific POS pair compositional functions. Positive values are blue, negative values red, and zero values are white. Certain phrasal relationships (e.g., DT-NN and VB-VB) exhibit headedness.}
	\label{fig:heatmaps}
\end{figure*}

The parameters $\bW$ in Eq.~\ref{eq:concat} and \ref{eq:tensor} can be estimated through standard linear regression techniques in conjunction with the data presented in \S\ref{sec:ppdb}.
These methods provide a natural way to regularize $\bW$ via $\ell_2$ (ridge) or $\ell_1$ (LASSO) regularization, which also helps handle noisy paraphrases.  
Parameters for the $\ell_1$-regularized concatenation model for select POS pairs are displayed in Fig.~\ref{fig:heatmaps}.\footnote{Parameters learned with $\ell_2$ regularization yield too many non-zero values, making visualization less informative.}
The heat-maps display the relative magnitude of parameters, with positive values colored blue, negative values colored red, and white cells indicating zero values. 
It is evident that the parameters learned from PPDB indicate a notion of linguistic headedness, namely that for particular POS pairs, the semantic information is primarily contained in the right word, but for others such as the noun--noun combination, each constituent's contribution is relatively more equal. 


\section{Measuring of Compositionality}
\label{sec:scoring}

The concatenation and tensor models compute an $N$-dimensional vector representation for a multi-word phrase by assuming the meaning of the phrase can be expressed in terms of the meaning of its constituents. 
This assumption holds true to varying degrees; while it clearly holds for ``large amount" and breaks down for ``cloud nine", it is partially valid for phrases such as ``zebra crossing" or ``crash course". 
In line with previous work, we assume a {\bf compositionality continuum} \cite{McCarthy2003}, but further conjecture that a phrase's level of compositionality is dependent on the specific context in which it occurs,  motivating a context-based approach (\S\ref{sec:context}) which scores compositionality by computing the likelihoods of surrounding context words given a phrase representation. 
The effect of context is directly measured through a comparison with context-independent methods from prior work \cite{Bannard2003,Reddy2011}

It is important to note that most prior work on compositionality scoring assumes access to \emph{both} word and phrase vector representations (for select phrases that will be evaluated) \emph{a priori}.  
The latter are distinct from representations that are computed from learned compositional functions as they are extracted directly from the corpus, which is an expensive procedure. 
Our aim is to develop compositional models that are applicable in downstream tasks, and thus assuming pre-existing phrase vectors is unreasonable.\footnote{If these phrase representations were easy to extract from corpora, that would obviate the need to learn compositional functions.}
Hence for phrases, we only rely on representations computed from our learned compositional functions. 

\subsection{At the Type Level}
\label{sec:independent}

Given vector representations for the constituent words in a phrase and the phrase itself, the idea behind the type-based model is to compute similarities between the constituent word representations and the phrasal representation, and average the similarities across the constituents. 
If the contexts in which a constituent word occurs, as dictated by its vector representation, are very different from the contexts of the composed phrase, as indicated by the cosine similarity between the word and phrase representations, then the phrase is likely to be non-compositional. 
Assuming unit-normalized word vectors $\bX, \bY$ and phrase vector $\bZ = \bF(\bX, \bY)$ computed from one of the learned models in \S\ref{sec:learning}:
\begin{align}
	g(\bX, \bY, \bZ) &= \alpha (\bX \cdot \bZ) + (1-\alpha)(\bY \cdot \bZ)
	\label{eq:cosine-sim}
\end{align}
where $\alpha$ is a hyperparameter that controls the contribution of individual constituents. 
This model leverages the average statistics computed over the training corpora (as encapsulated in the word and phrase vectors) to detect compositionality, and is the primary way compositionality has been evaluated previously \cite{Reddy2011,Kiela2013}. 
Note that for the simple additive model $\bF(\bX, \bY) = \bX + \bY$ with unit-normalized word vectors, $g(\bX, \bY, \bZ)$ is independent of $\alpha$.

\subsection{At the Token Level}
\label{sec:context}

Eq.~\ref{eq:cosine-sim} scores phrases for compositionality regardless of the context that these phrases occur in. 
However, phrases such as ``big fish" or ``heavy metal" may occur in both compositional and non-compositional situations, depending on the nature and topic of the texts they occur in.\footnote{In fact, human annotators have access to such context when making compositionality judgments.}
Here, we propose a context-driven model for compositionality detection, inspired by the skip-gram model for learning word representations \cite{Mikolov2013b}. 
The intuition is simple: if a phrase is compositional, it should be sufficiently predictive of the context words around it; otherwise, it is acting in a non-compositional manner. 
Thus, we would like to compute the likelihood of the context ($\boldsymbol{c}$) given a phrasal representation ($\bZ = \bF(\bX, \bY)$) and normalization constant $Z$:
\begin{align}
	P(\boldsymbol{c} \mid \bX,\bY) = \prod_{i=1}^{|\boldsymbol{c}|} \frac{\exp \bF(\bX,\bY) \cdot \boldsymbol{v}_{c_i}}{Z(\bX,\bY)}.
	\label{eq:likelihood}
\end{align}
As explained in \newcite{Goldberg2014}, the context representations are distinct from the word representations.
In practice, we compute the log-likelihood averaged over the context words or the perplexity instead of the actual likelihood. 

\section{Evaluation}
\label{sec:experiments}

\begin{figure*}[t!]
	\begin{center}
	\begin{subfigure}{\columnwidth}
		\centering
		\includegraphics[width=\columnwidth,keepaspectratio=true]{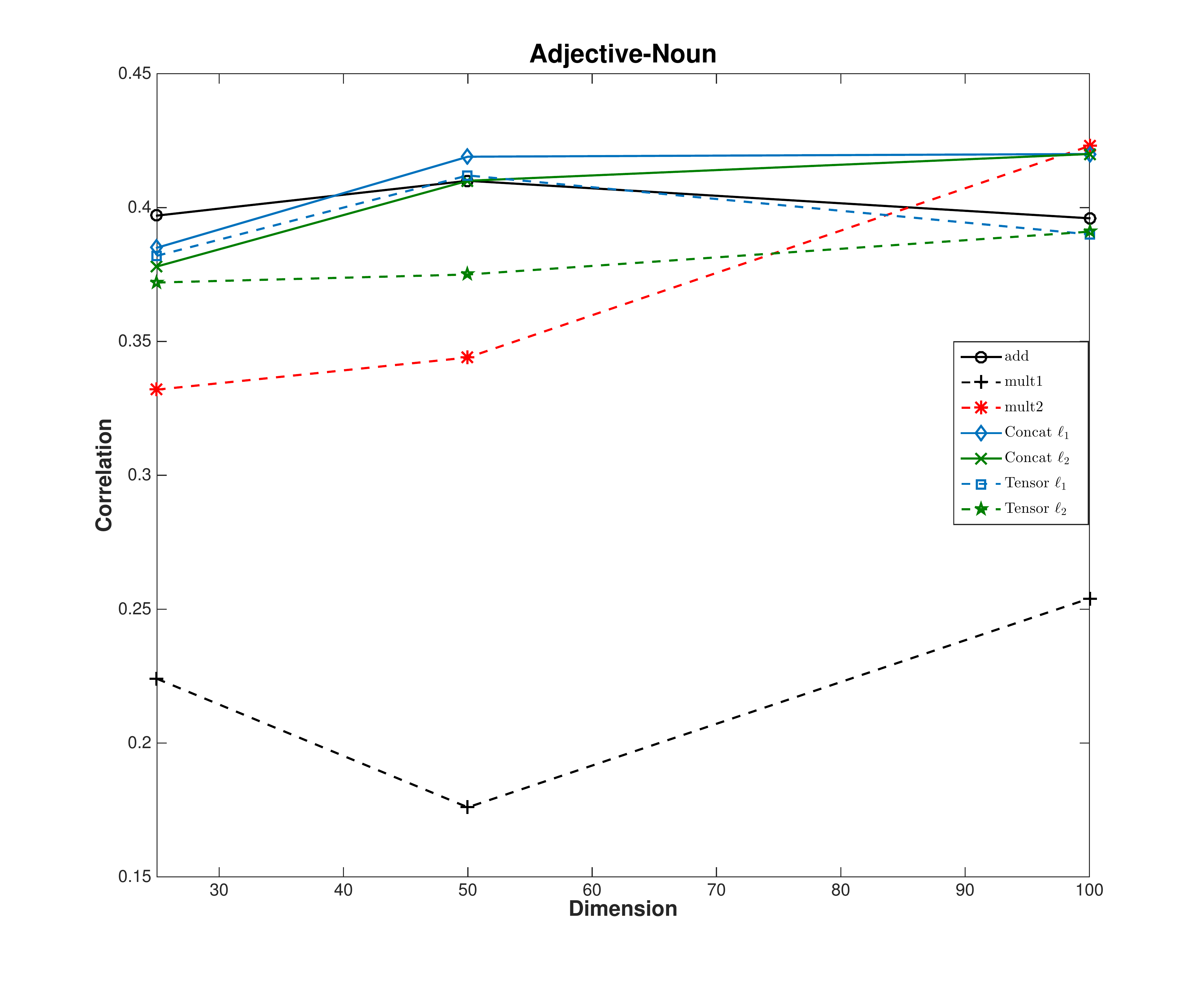}	
		\caption{\small JJ--NN}
		\label{fig:jj_nn_result}			
	\end{subfigure}
	\begin{subfigure}{\columnwidth}
		\centering
		\includegraphics[width=\columnwidth,keepaspectratio=true]{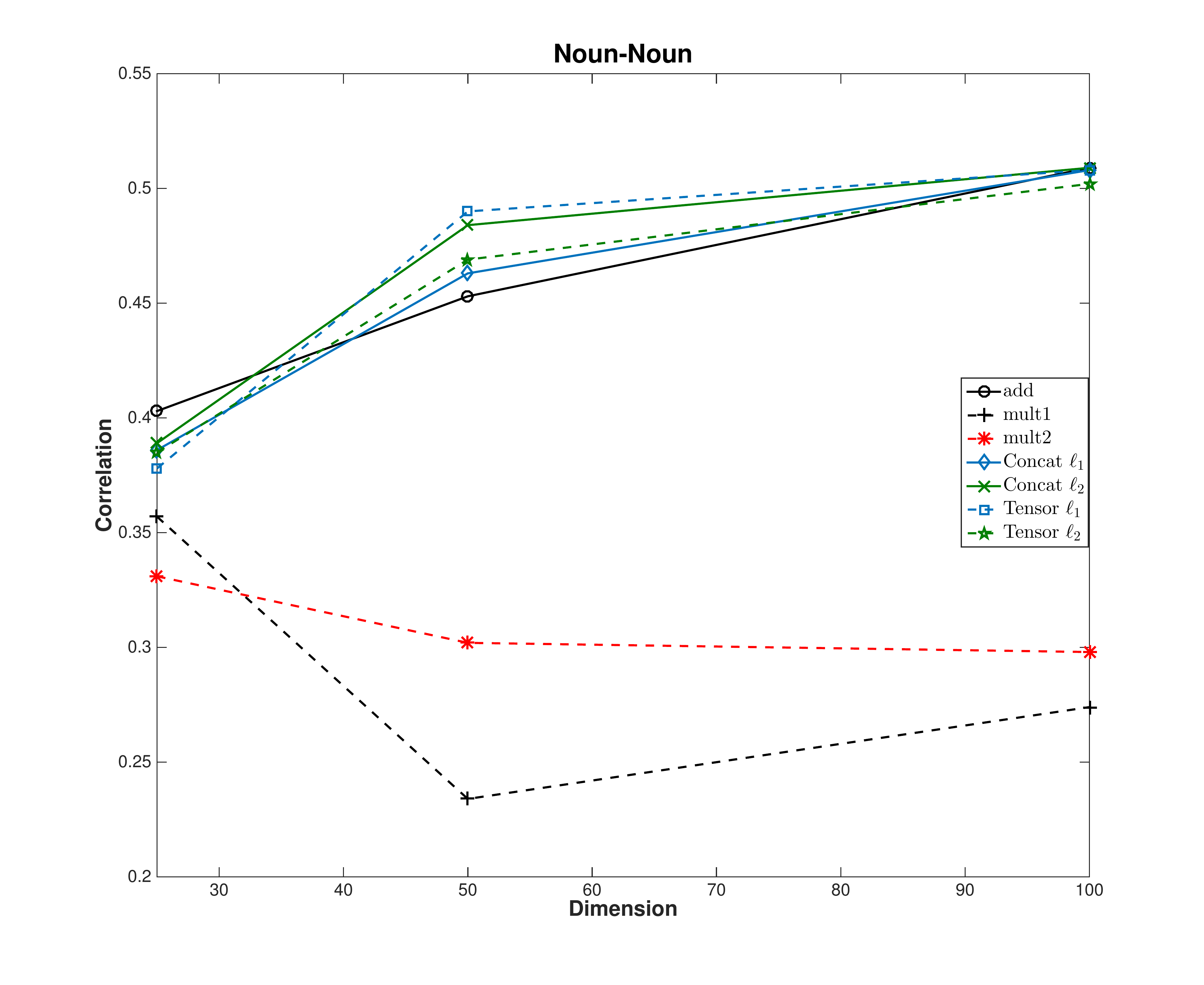}		
		\caption {\small NN--NN}
		\label{fig:nn_nn_result}
	\end{subfigure}
	\end{center}
	\caption{Spearman's $\rho$ correlation with respect to human judgments for the adjective--noun and noun--noun phrase similarity tasks. Dashed lines correspond to tensor models (and baselines), and solid lines are concatenation models (and additive baseline).}
	\label{fig:correlation}
\end{figure*}

Our experiments had three aims: first, demonstrate that the compositional functions learned using paraphrase supervision compute semantically meaningful results for compositional phrases by evaluating on a phrase similarity task (\S\ref{sec:phrasesim-eval}); second, verify the hypothesis that compositionality is context-dependent by comparing a type-based and token-based approach on a compound noun evaluation task (\S\ref{sec:compo-eval}); and third, determine if the compositionality-scoring models based on learned representations improve the translations produced by a state-of-the-art phrase-based MT system (\S\ref{sec:mt-eval}).  

The word vectors used in all of our experiments were produced by \texttt{word2vec}\footnote{\texttt{http://code.google.com/p/word2vec}} using the skip-gram model with 20 negative samples, a context window size of 10, a minimum token count of 3, and sub-sampling of frequent words with a parameter of $10^{-5}$.  
We extracted corpus statistics for \texttt{word2vec} using the AFP portion of the English Gigaword\footnote{LDC2011T07}, which consists of 887.5 million tokens. 
The code used to generate the results is available at \texttt{http://www.github.com/xyz}, and the evaluation datasets are publicly available. 

\subsection{Phrasal Similarity}
\label{sec:phrasesim-eval}

For the phrase similarity task we first compare our concatenation and tensor models learned using $\ell_1$ and $\ell_2$ regularization to three baselines:
\begin{itemize}[noitemsep]
	\item \textsc{add}: $\bF(\bX, \bY) = \bX + \bY$
	\item \textsc{mult1}: $f_i(x_i, y_i) = x_i y_i$
	\item \textsc{mult2}: $f_i(x_i, y_i) = \alpha_i x_i y_i$
\end{itemize}
Other additive models from previous work \cite{Mitchell2010,Zanzotto2010,Blacoe2012} that impose varying amounts of structural assumptions on the semantic interactions between word representations e.g., $f_i(x_i, y_i) = \alpha_i x_i + \beta_i y_i$ or $\bF(\bX, \bY) = \alpha\bX + \beta\bY$ are subsumed by our concatenation model. 
The regularization strength hyperparameter for $\ell_1$ and $\ell_2$ regularization was selected using 5-fold cross-validation on the PPDB training data. 

We evaluated the phrase compositionality models on the adjective--noun and noun--noun phrase similarity tasks compiled by \newcite{Mitchell2010}, using the same evaluation scheme as in the original work.\footnote{The evaluation set also consists of verb-object phrases constructed from dependency relations and their similarity, but such phrases generally do not fall into our phrasal definition since the words are not contiguous.}
Spearman's $\rho$ between phrasal similarities derived from our compositional functions and the human annotators (computed individually per annotator and then averaged across all annotators) was the evaluation measure. 

Figure \ref{fig:correlation} presents the correlation results for the two POS pair types as a function of the dimensionality $N$ of the representations for the concatenation models (and additive baseline) and tensor models (and multiplicative baselines). 
The concatenation models seem more effective than the tensor models in the adjective--noun case and give roughly the same performance on the noun--noun dataset, which is consistent with previous work that uses dense, low-dimensional representations \cite{Guevara2011,Hermann2013,Hashimoto2014}.\footnote{Multiplicative or tensor-based models seem to do better on sparse, high-dimensional representations \cite{Mitchell2010,Baroni2010} since multiplication represents a conjunction of co-occurrence features.}
Since the concatenation model involve fewer parameters, we use it as the compositional model of choice for subsequent experiments. 
The absolute results are also consistent with state-of-the-art results on this dataset \cite{Blacoe2012,Hashimoto2014}\footnote{There are differences in the corpora and experimental setup which explains the small discrepancies.}, indicating that paraphrases are an excellent source of information for learning compositional functions and a reasonable alternative to human-annotated training sets. 
For reference, the inter-annotator agreements are 0.52 for the adjective--noun evaluation and 0.51 for the noun--noun one. 
The unweighted additive baseline is surprisingly very strong on the noun--noun set, so we also compare against it in subsequent experiments. 

\subsection{Compositionality}
\label{sec:compo-eval}

To evaluate the compositionality-scoring models, we used the compound noun compositionality dataset introduced in \newcite{Reddy2011}.  
This dataset consists of 2670 annotations of 90 compound-noun phrases exhibiting varying levels compositionality, with scores ranging from 0 to 5 provided by 30 annotators. 
It also contains three to five example sentences of these phrases that were shown to the annotators, which we make use of in our context-dependent model. 
Consistent with the original work, Spearman's $\rho$ is computed on the averaged compositionality score for a phrase across all the annotators that scored that phrase (which varies per phrase). 
For computing the compositional functions, we evaluate three of the best performing setups from \S\ref{sec:phrasesim-eval}: the $\ell_1$ and $\ell_2$-regularized concatenation models, and the simple additive baseline. 

\begin{table}[h!]
	\small
  \begin{center}
    \begin{tabular}{p{0.32\linewidth}ccc}
      \hline
      & \multicolumn{3}{c}{$\bm{\rho}$} \\
	  {\bf Setup} & {\bf \textsc{Add}} & {\bf Concat. $\ell_1$} & {\bf Concat. $\ell_2$} \\
	  \hline
      Type $\alpha=0.25$ & $\mid$ & 0.43 & 0.46 \\
	  Type $\alpha=0.5$ &  0.41 & 0.42 & 0.47 \\
	  Type $\alpha=0.75$ & $\mid$ & 0.41 & 0.43 \\
	  \hline 
	  Token $l=4$ & {\bf 0.55} & {\bf 0.58} & 0.58 \\
      Token $l=6$ & 0.54 & 0.57 & {\bf 0.59} \\
	  Token $l=8$  & 0.53 & 0.58 & 0.59 \\
	\end{tabular}
  \end{center}
  \caption{Correlation between model judgments on phrase compositionality and human judgments, measured by Spearman's $\rho$. Context-dependent (token-based) and concatenation models do better.}
  \label{tab:comp-results}
\end{table}

For the context-independent model, we select the hyperparameter $\alpha$ in Eq.~\ref{eq:cosine-sim} from the values $\{0.25, 0.5, 0.75\}$. 
For the context-dependent model, we vary the context window size $|\boldsymbol{c}|$ by selecting from the values $\{4, 6, 8\}$. 
Table \ref{tab:comp-results} presents Spearman's $\rho$ for these setups. 
In all cases, the context-dependent models outperform the context-independent ones, and using a relatively simple token-based model we can approximately match the performance of the Bayesian model proposed by \newcite{Hermann2012}.  
The concatenation models are also consistently better than the additive compositional model, indicating the benefit of learning the compositional parameters via PPDB. 

\subsection{Machine Translation}
\label{sec:mt-eval}

While any truly successful model of semantics must match human intuitions, understanding the applications of our models is likewise important. 
To this end, we consider the problem of machine translation, operating under the hypothesis that sentences which express their meaning non-compositionally should also translate non-compositionally. 

Modern phrase-based translation systems are faced with a large number of possible segmentations of a source-language sentence during decoding, and all segmentations are considered equally likely \cite{Koehn2003}.  
Thus, it would be helpful to provide guidance on more likely segmentations, as dictated by the compositionality scores of the phrases extracted from a sentence, to the decoder. 
A low compositionality score would ideally force the decoder to consider the entire phrase as a translation unit, due to its unique semantic characteristics.
Correspondingly, a high score informs the decoder that it is safe to rely on word-level translations of the phrasal constituents. 
Thus, if we reveal to the translation system that a phrase is non-compositional, it should be able to learn that translation decisions which translate it as a unit are to be favored, leading to better translations.


To test this hypothesis, we built an English-Spanish MT system using the \textsc{cdec} decoder \cite{Dyer2010} for the entire training pipeline (word alignments, phrase extraction, feature weight tuning, and decoding).
Corpora from the WMT 2011 evaluation\footnote{\texttt{http://www.statmt.org/wmt11/}} was used to build the translation and language models, and for tuning (on \texttt{news-test2010}) and evaluation (on \texttt{news-test2011}), with scoring done using BLEU \cite{Papineni2002}. 
The baseline is a hierarchical phrase-based system \cite{Chiang2007} with a 4-gram language model, with feature weights tuned using MIRA \cite{Chiang2012}. 
For features, each translation rule is decorated with two lexical and phrasal features corresponding to the forward $(e|f)$ and backward $(f|e)$ conditional log frequencies, along with the log joint frequency $(e,f)$, the log frequency of the source phrase $(f)$, and whether the phrase pair or the source phrase is a singleton. 
Weights for the language model, glue rule, and word penalty are also tuned. 
This setup (Baseline) achieves scores \emph{en par} with the published WMT results. 

We added the compositionality score as an additional feature, and also added two binary-valued features: the first indicates if the given translation rule has not been decorated with a compositionality score (either because it consists of non-terminals only or the lexical items in the translation rule are unigrams), and correspondingly the second feature indicates if the translation rule has been scored. 
Therefore, an appropriate additional baseline would be to mark translation rules with these indicator functions but without the scores, akin to identifying rules with phrases in them (Baseline + SegOn). 

\begin{table}[h!]
  \begin{center}
    \begin{tabular}{p{0.4\linewidth}rr}
      \hline
	  & \multicolumn{2}{c}{\bf BLEU} \\
      Setup &  \multicolumn{1}{c}{Dev} & \multicolumn{1}{c}{Test} \\
	  \hline
	  Baseline & 25.23 (0.05) & 26.89 (0.13) \\
      Baseline + SegOn & 25.15 (0.21) & 26.87 (0.19) \\
	  $\ell_2$ CosSim $\alpha=0.5$ &  25.08 (0.03) & {\bf26.99} (0.04) \\
  	  \textsc{Add} $l=4$ &  24.85 (0.12) & 26.82 (0.05) \\
      $\ell_1$ $l=4$ &  25.08 (0.08) & {\bf 27.03} (0.10) \\
	  $\ell_2$  $l=6$ & 25.12 (0.22) & {\bf 27.26} (0.21) \\
	\end{tabular}
  \end{center}
  \caption{MT results. Bold results are statistically significant, and our best context-dependent setup is 0.4 BLEU points better than the baseline.}
  \label{tab:mt-results}
\end{table}

Table \ref{tab:mt-results} presents the results of the MT evaluation, comparing the baselines to the best-performing context-independent and dependent scoring models from \S\ref{sec:compo-eval}. 
The scores have been averaged over three tuning runs with standard deviation in parentheses; bold results on the test set are statistically significant ($p < 0.05$) with respect to the baseline. 
While knowledge of relative compositionality consistently helps, the improvements using the context-dependent scoring models, especially with the $\ell_2$ concatenation model, are noticeably better. 

\section{Related Work}

There has been a large amount of work on compositional models that operate on vector representations of words. 
With some exceptions \cite{Mitchell2008,Mitchell2010}, all of these approaches are lexicalized i.e., parameters (generally in the form of vectors, matrices, or tensors) for \emph{specific} words are learned, which works well for frequently occurring words but fails when dealing with compositions of arbitrary word sequences containing infrequent words. 
The functions are either learned with a neural network architecture \cite[\emph{inter alia}]{Socher2013} or as a linear regression \cite{Baroni2010}; the latter require phrase representations extracted directly from the corpus for supervision, which can be computationally expensive and statistically inefficient. 
In contrast, we obtain this information through many-to-one PPDB mappings. 
Most of these models also require additional syntactic \cite{Socher2012} or semantic \cite{Hermann2013,Grefenstette2013} resources; on the other hand, our proposed approach only requires a shallow syntactic parse (POS tags).  
Recent efforts to make these models more practical \cite{Paperno2014} attempt to reduce their statistically complex and overly-parametrized nature, but with the exception of \newcite{Zanzotto2010}, who propose a way to extract compositional function training examples from a dictionary, these models generally require human-annotated data to work.  

Most models that score the relative (non-) compositionality of phrases do so in a context-independent manner. 
A central idea is to replace phrase constituents with semantically-related words and compute the similarity of the new phrase to the original \cite{Kiela2013,Salehi2014} or make use of a variety of lexical association measures \cite{Lin1999,Pecina2006}. 
\newcite{Sporleder2009} however, do make use of context in a token-based approach, where the context in which a phrase occurs as well as the phrase itself is modeled as a lexical chain, and the cohesion of the chain is measured as an indicator of a phrase's compositionality. 
Cohesion is computed using a web search engine-based measure, whereas we use a probabilistic model of context given a phrase representation. 
\newcite{Hermann2012} propose a Bayesian generative model that is also context-based, but learning and inference is done through a relatively expensive Gibbs sampling scheme.  

In the context of MT, \newcite{Zhang2008b} present a Bayesian model that learns non-compositional phrases from a synchronous parse tree of a sentence pair.
However, the primary aim of their work is phrase extraction for MT, and the non-compositional constraints are only applied to make the space of phrase pairs more tractable when bootstrapping their phrasal parser from their word-based parser. 
In contrast, we score every phrase that is extracted with the standard phrase extraction heuristics \cite{Chiang2007}, allowing the decoder to make the final decision on the impact of compositionality scores in translation. 
Thus, our work is more similar to \newcite{Xiong2010}, who propose maximum entropy classifiers that mark positions between words in a sentence as being a phrase boundary or not, and integrate these scores as additional features in an MT system.  

\section{Conclusion}

In this work, we presented two new sources of information for compositionality modeling and scoring, paraphrase information and context. 
For modeling, we showed that the paraphrase-learned compositional representations performs as well on a phrase similarity task as the average human annotator.
For scoring, the importance of context was shown through the comparison of context-independent and dependent models. 
Improvements by the context-dependent model on an extrinsic machine translation task corroborate the utility of these additional knowledge sources. 
We hope that this work encourages further research in making compositional semantic approaches applicable in downstream tasks. 

\bibliographystyle{acl}
\bibliography{bibliography}

\begin{thebibliography}{}

\bibitem[\protect\citename{Bannard and Callison-Burch}2005]{Bannard2005}
Colin Bannard and Chris Callison-Burch.
\newblock 2005.
\newblock Paraphrasing with bilingual parallel corpora.
\newblock In {\em Proceedings of ACL}.

\bibitem[\protect\citename{Bannard \bgroup et al.\egroup }2003]{Bannard2003}
Colin Bannard, Timothy Baldwin, and Alex Lascarides.
\newblock 2003.
\newblock A statistical approach to the semantics of verb-particles.
\newblock In {\em Proceedings of the ACL Workshop on Multiword Expressions:
  Analysis, Acquisition and Treatment}.

\bibitem[\protect\citename{Baroni and Zamparelli}2010]{Baroni2010}
Marco Baroni and Roberto Zamparelli.
\newblock 2010.
\newblock Nouns are vectors, adjectives are matrices: Representing
  adjective-noun constructions in semantic space.
\newblock In {\em Proceedings of EMNLP}.

\bibitem[\protect\citename{Blacoe and Lapata}2012]{Blacoe2012}
William Blacoe and Mirella Lapata.
\newblock 2012.
\newblock A comparison of vector-based representations for semantic
  composition.
\newblock In {\em Proceedings of EMNLP-CoNLL}.

\bibitem[\protect\citename{Chiang}2007]{Chiang2007}
David Chiang.
\newblock 2007.
\newblock Hierarchical phrase-based translation.
\newblock {\em Computational Linguistics}.

\bibitem[\protect\citename{Chiang}2012]{Chiang2012}
David Chiang.
\newblock 2012.
\newblock Hope and fear for discriminative training of statistical translation
  models.
\newblock {\em Journal of Machine Learning Research}.

\bibitem[\protect\citename{Collobert \bgroup et al.\egroup
  }2011]{Collobert2011}
Ronan Collobert, Jason Weston, L\'{e}on Bottou, Michael Karlen, Koray
  Kavukcuoglu, and Pavel Kuksa.
\newblock 2011.
\newblock Natural language processing (almost) from scratch.
\newblock {\em Journal of Machine Learning Research}.

\bibitem[\protect\citename{Dyer \bgroup et al.\egroup }2010]{Dyer2010}
Chris Dyer, Adam Lopez, Juri Ganitkevitch, Johnathan Weese, Ferhan Ture, Phil
  Blunsom, Hendra Setiawan, Vladimir Eidelman, and Philip Resnik.
\newblock 2010.
\newblock cdec: A decoder, alignment, and learning framework for finite-state
  and context-free translation models.
\newblock In {\em Proceedings of ACL}.

\bibitem[\protect\citename{Ganitkevitch and
  Callison-Burch}2014]{Ganitkevich2014}
Juri Ganitkevitch and Chris Callison-Burch.
\newblock 2014.
\newblock The multilingual paraphrase database.
\newblock In {\em Proceedings of LREC}.

\bibitem[\protect\citename{Ganitkevitch \bgroup et al.\egroup
  }2013]{Ganitkevich2013}
Juri Ganitkevitch, Benjamin {Van Durme}, and Chris Callison-Burch.
\newblock 2013.
\newblock {PPDB}: The paraphrase database.
\newblock In {\em Proceedings of NAACL-HLT}.

\bibitem[\protect\citename{Goldberg and Levy}2014]{Goldberg2014}
Yoav Goldberg and Omer Levy.
\newblock 2014.
\newblock word2vec explained: deriving mikolov et al.'s negative-sampling
  word-embedding method.
\newblock {\em CoRR: abs/1402.3722}.

\bibitem[\protect\citename{Grefenstette \bgroup et al.\egroup
  }2013]{Grefenstette2013}
Edward Grefenstette, Georgiana Dinu, Yao-Zhang Zhang, Mehrnoosh Sadrzadeh, and
  Marco Baroni.
\newblock 2013.
\newblock Multi-step regression learning for compositional distributional
  semantics.
\newblock In {\em Proceedings of IWCS}.

\bibitem[\protect\citename{Guevara}2011]{Guevara2011}
Emiliano Guevara.
\newblock 2011.
\newblock Computing semantic compositionality in distributional semantics.
\newblock In {\em Proceedings of IWCS}.

\bibitem[\protect\citename{Hashimoto \bgroup et al.\egroup
  }2014]{Hashimoto2014}
Kazuma Hashimoto, Pontus Stenetorp, Makoto Miwa, and Yoshimasa Tsuruoka.
\newblock 2014.
\newblock Jointly learning word representations and composition functions using
  predicate-argument structures.
\newblock In {\em Proceedings of EMNLP}.

\bibitem[\protect\citename{Hermann and Blunsom}2013]{Hermann2013}
Karl~Moritz Hermann and Phil Blunsom.
\newblock 2013.
\newblock The role of syntax in vector space models of compositional semantics.
\newblock In {\em Proceedings of ACL}.

\bibitem[\protect\citename{Hermann \bgroup et al.\egroup }2012]{Hermann2012}
Karl~Moritz Hermann, Phil Blunsom, and Stephen Pulman.
\newblock 2012.
\newblock An unsupervised ranking model for noun-noun compositionality.
\newblock In {\em Proceedings of {*SEM}}.

\bibitem[\protect\citename{Kalchbrenner \bgroup et al.\egroup
  }2014]{Kalchbrenner2014}
Nal Kalchbrenner, Edward Grefenstette, and Phil Blunsom.
\newblock 2014.
\newblock A convolutional neural network for modelling sentences.
\newblock In {\em Proceedings of ACL}.

\bibitem[\protect\citename{Kiela and Clark}2013]{Kiela2013}
Douwe Kiela and Stephen Clark.
\newblock 2013.
\newblock Detecting compositionality of multi-word expressions using nearest
  neighbours in vector space models.
\newblock In {\em Proceedings of EMNLP}.

\bibitem[\protect\citename{Koehn \bgroup et al.\egroup }2003]{Koehn2003}
Philipp Koehn, Franz~Josef Och, and Daniel Marcu.
\newblock 2003.
\newblock Statistical phrase-based translation.
\newblock In {\em Proceedings of NAACL}.

\bibitem[\protect\citename{Le and Mikolov}2014]{Le2014}
Quoc Le and Tomas Mikolov.
\newblock 2014.
\newblock Distributed representations of sentences and documents.
\newblock In {\em Proceedings of ICML}.

\bibitem[\protect\citename{Lin}1999]{Lin1999}
Dekang Lin.
\newblock 1999.
\newblock Automatic identification of non-compositional phrases.
\newblock In {\em Proceedings of ACL}.

\bibitem[\protect\citename{McCarthy \bgroup et al.\egroup }2003]{McCarthy2003}
Diana McCarthy, Bill Keller, and John Carroll.
\newblock 2003.
\newblock Detecting a continuum of compositionality in phrasal verbs.
\newblock In {\em Proceedings of the ACL Workshop on Multiword Expressions:
  Analysis, Acquisition and Treatment}.

\bibitem[\protect\citename{Mikolov \bgroup et al.\egroup }2013]{Mikolov2013b}
Tomas Mikolov, Ilya Sutskever, Kai Chen, Greg Corrado, and Jeffrey Dean.
\newblock 2013.
\newblock Distributed representations of words and phrases and their
  compositionality.
\newblock {\em CoRR: abs/1310.4546}.

\bibitem[\protect\citename{Mitchell and Lapata}2008]{Mitchell2008}
Jeff Mitchell and Mirella Lapata.
\newblock 2008.
\newblock Vector-based models of semantic composition.
\newblock In {\em Proceedings of ACL-HLT}.

\bibitem[\protect\citename{Mitchell and Lapata}2010]{Mitchell2010}
Jeff Mitchell and Mirella Lapata.
\newblock 2010.
\newblock Composition in distributional models of semantics.
\newblock {\em Cognitive Science}.

\bibitem[\protect\citename{Paperno \bgroup et al.\egroup }2014]{Paperno2014}
Denis Paperno, Nghia~The Pham, and Marco Baroni.
\newblock 2014.
\newblock A practical and linguistically-motivated approach to compositional
  distributional semantics.
\newblock In {\em Proceedings of ACL}.

\bibitem[\protect\citename{Papineni \bgroup et al.\egroup }2002]{Papineni2002}
Kishore Papineni, Salim Roukos, Todd Ward, and Wei-Jing Zhu.
\newblock 2002.
\newblock {BLEU}: A method for automatic evaluation of machine translation.
\newblock In {\em Proceedings of ACL}.

\bibitem[\protect\citename{Pecina and Schlesinger}2006]{Pecina2006}
Pavel Pecina and Pavel Schlesinger.
\newblock 2006.
\newblock Combining association measures for collocation extraction.
\newblock In {\em Proceedings of COLING-ACL}.

\bibitem[\protect\citename{Reddy \bgroup et al.\egroup }2011]{Reddy2011}
Siva Reddy, Diana McCarthy, and Suresh Manandhar.
\newblock 2011.
\newblock An empirical study on compositionality in compound nouns.
\newblock In {\em Proceedings of IJCNLP}.

\bibitem[\protect\citename{Sag \bgroup et al.\egroup }2002]{Sag2002}
Ivan~A. Sag, Timothy Baldwin, Francis Bond, Ann~A. Copestake, and Dan
  Flickinger.
\newblock 2002.
\newblock Multiword expressions: A pain in the neck for nlp.
\newblock In {\em Proceddings of CICLing}.

\bibitem[\protect\citename{Sahlgren}2006]{Sahlgren2006}
Magnus Sahlgren.
\newblock 2006.
\newblock {\em The Word-Space Model: Using distributional analysis to represent
  syntagmatic and paradigmatic relations between words in high-dimensional
  vector spaces}.
\newblock {Ph.D.} thesis, Department of Linguistics, Stockholm University.

\bibitem[\protect\citename{Salehi \bgroup et al.\egroup }2014]{Salehi2014}
Bahar Salehi, Paul Cook, and Timothy Baldwin.
\newblock 2014.
\newblock Using distributional similarity of multi-way translations to predict
  multiword expression compositionality.
\newblock In {\em Proceedings of EACL}.

\bibitem[\protect\citename{Socher \bgroup et al.\egroup }2011]{Socher2011}
Richard Socher, Jeffrey Pennington, Eric~H. Huang, Andrew~Y. Ng, and
  Christopher~D. Manning.
\newblock 2011.
\newblock {Semi-Supervised Recursive Autoencoders for Predicting Sentiment
  Distributions}.
\newblock In {\em Proceedings of EMNLP}.

\bibitem[\protect\citename{Socher \bgroup et al.\egroup }2012]{Socher2012}
Richard Socher, Brody Huval, Christopher~D. Manning, and Andrew~Y. Ng.
\newblock 2012.
\newblock {Semantic Compositionality Through Recursive Matrix-Vector Spaces}.
\newblock In {\em Proceedings EMNLP}.

\bibitem[\protect\citename{Socher \bgroup et al.\egroup }2013]{Socher2013}
Richard Socher, Alex Perelygin, Jean Wu, Jason Chuang, Christopher~D. Manning,
  Andrew Ng, and Christopher Potts.
\newblock 2013.
\newblock Recursive deep models for semantic compositionality over a sentiment
  treebank.
\newblock In {\em Proceedings of EMNLP}.

\bibitem[\protect\citename{Sporleder and Li}2009]{Sporleder2009}
Caroline Sporleder and Linlin Li.
\newblock 2009.
\newblock Unsupervised recognition of literal and non-literal use of idiomatic
  expressions.
\newblock In {\em Proceedings of EACL}.

\bibitem[\protect\citename{Turian \bgroup et al.\egroup }2010]{Turian2010}
Joseph Turian, Lev Ratinov, and Yoshua Bengio.
\newblock 2010.
\newblock Word representations: A simple and general method for semi-supervised
  learning.
\newblock In {\em Proceedings of ACL}.

\bibitem[\protect\citename{Turney and Pantel}2010]{Turney2010}
Peter~D. Turney and Patrick Pantel.
\newblock 2010.
\newblock From frequency to meaning: Vector space models of semantics.
\newblock {\em Journal of Artificial Intelligence Research}.

\bibitem[\protect\citename{Xiong \bgroup et al.\egroup }2010]{Xiong2010}
Deyi Xiong, Min Zhang, and Haizhou Li.
\newblock 2010.
\newblock Learning translation boundaries for phrase-based decoding.
\newblock In {\em Proceedings of NAACL-HLT}.

\bibitem[\protect\citename{Zanzotto \bgroup et al.\egroup }2010]{Zanzotto2010}
Fabio~Massimo Zanzotto, Ioannis Korkontzelos, Francesca Fallucchi, and Suresh
  Manandhar.
\newblock 2010.
\newblock Estimating linear models for compositional distributional semantics.
\newblock In {\em Proceedings of COLING}.

\bibitem[\protect\citename{Zhang \bgroup et al.\egroup }2008]{Zhang2008b}
Hao Zhang, Chris Quirk, Robert~C. Moore, and Daniel Gildea.
\newblock 2008.
\newblock {Bayesian} learning of non-compositional phrases with synchronous
  parsing.
\newblock In {\em Proceedings of ACL-HLT}.

\end{thebibliography}
\end{document}